%% file: main.tex
\documentclass{article}

\usepackage{PRIMEarxiv}

\usepackage[utf8]{inputenc} 
\usepackage[T1]{fontenc}    
\usepackage{hyperref}       
\usepackage{url}            
\usepackage{booktabs}       
\usepackage{amsfonts}       
\usepackage{nicefrac}       
\usepackage{microtype}      
\usepackage{lipsum}
\usepackage{fancyhdr}       
\usepackage{graphicx}       
\graphicspath{{media/}}     

\usepackage{hyperref}
\usepackage{url}
\usepackage{caption}
\usepackage{multirow}
\usepackage{booktabs}
\usepackage{amsmath}
\usepackage{enumitem}

\newif\ifshowdocvqatest
\showdocvqatesttrue  

\newcommand{\conditionaltext}[2]{%
  \ifshowdocvqatest
    #1 
  \else
    #2 
  \fi
}

\title{SDRT: Enhance Vision-Language Models by Self-Distillation with Diverse Reasoning Traces
}

\author{Guande Wu\textsuperscript{1}\thanks{Work was done during an internship at at AWS AI.}, Huan Song\textsuperscript{2}, Yawei Wang\textsuperscript{2}, Qiaojing Yan\textsuperscript{2}, Yijun Tian\textsuperscript{2}, Lin Lee Cheong\textsuperscript{2}, Panpan Xu\textsuperscript{2}\\
\textsuperscript{1}New York University, \textsuperscript{2} AWS AI \\
{\tt\small guandewu@nyu.edu, \{huanso, yawenwan, qiaojiny, yijunt, lcheong, xupanpan\}@amazon.com, 
}
}

\begin{document}
\maketitle



\begin{abstract}

\input{meta/abstract}

\end{abstract}


\input{sections/01_intro}
\input{sections/05_related}

\input{sections/02b_framework_hs}
\input{sections/04_experiments}

\input{sections/06_limitations}
\input{sections/07_conclusion}


\bibliographystyle{unsrt}  
\bibliography{main}

\end{document}


\maketitle

\newif\ifshowdocvqatest
\showdocvqatesttrue  

\newcommand{\conditionaltext}[2]{%
  \ifshowdocvqatest
    #1 
  \else
    #2 
  \fi
}






\input{supp/prompt}


%% file: meta/abstract.tex
Reasoning is increasingly crucial for various tasks. While chain-of-thought prompting enables large language models to leverage reasoning effectively, harnessing the reasoning capabilities of Vision-Language Models (VLMs) remains challenging. To solve this problem, we propose a novel self-distillation framework that enhances the reasoning capabilities of the model. The proposed framework introduces several key innovations. We start by employing a prompt library tailored to visual reasoning tasks to generate diverse in-context questions and utilize a two-step reasoning procedure to derive reasoning-guided responses. These responses are then used for self-distillation, enabling the model to internalize the reasoning process. Additionally, we improve the model architecture with several innovative components, including an intervention adapter for efficient parameter updates, a cross-modal skip connection to facilitate information exchange between modalities, and an ensemble learning algorithm to integrate diverse reasoning from multiple in-context questions. Extensive experiments show that our method significantly improves the baseline performance across five VQA datasets.

%% file: sections/01_intro.tex
\section{Introduction}


The rapid advancements of large language models (LLMs) have significantly accelerated progress in natural language processing. Concurrently, Multi-modal LLMs, or Vision-Language Models (VLMs), opens new frontiers for addressing complex vision-language tasks such as visual question answering (VQA). These tasks inherently demand sophisticated multi-modal reasoning capabilities, requiring models to interpret and integrate rich visual content with textual information seamlessly. 

Such design of the VLMs, however, poses challenges in harnessing their reasoning capability, which is constrained by the abilities of pre-trained models.
Existing LLMs rely on chain-of-thought (CoT) prompting, a key method for facilitating advanced reasoning~\cite{DBLP:conf/nips/Wei0SBIXCLZ22, chu-etal-2024-navigate}. However, it has been observed that it is difficult to directly apply CoT on VLM for cross-modal reasoning. Various prompting or model fine-tuning methods have been developed to mitigate this~\cite{ DBLP:journals/ijcv/GaoGZMFZLQ24, wu2023role, zhang2024vision, DBLP:conf/eccv/JiaTCCBHL22}. Still, significant challenges remain in leveraging the reasoning capabilities of VLMs, which requires reasoning across both modalities to effectively integrate and analyze visual and textual data. 


\begin{figure}[tp]
    \centering
            \includegraphics[width=0.5\textwidth]{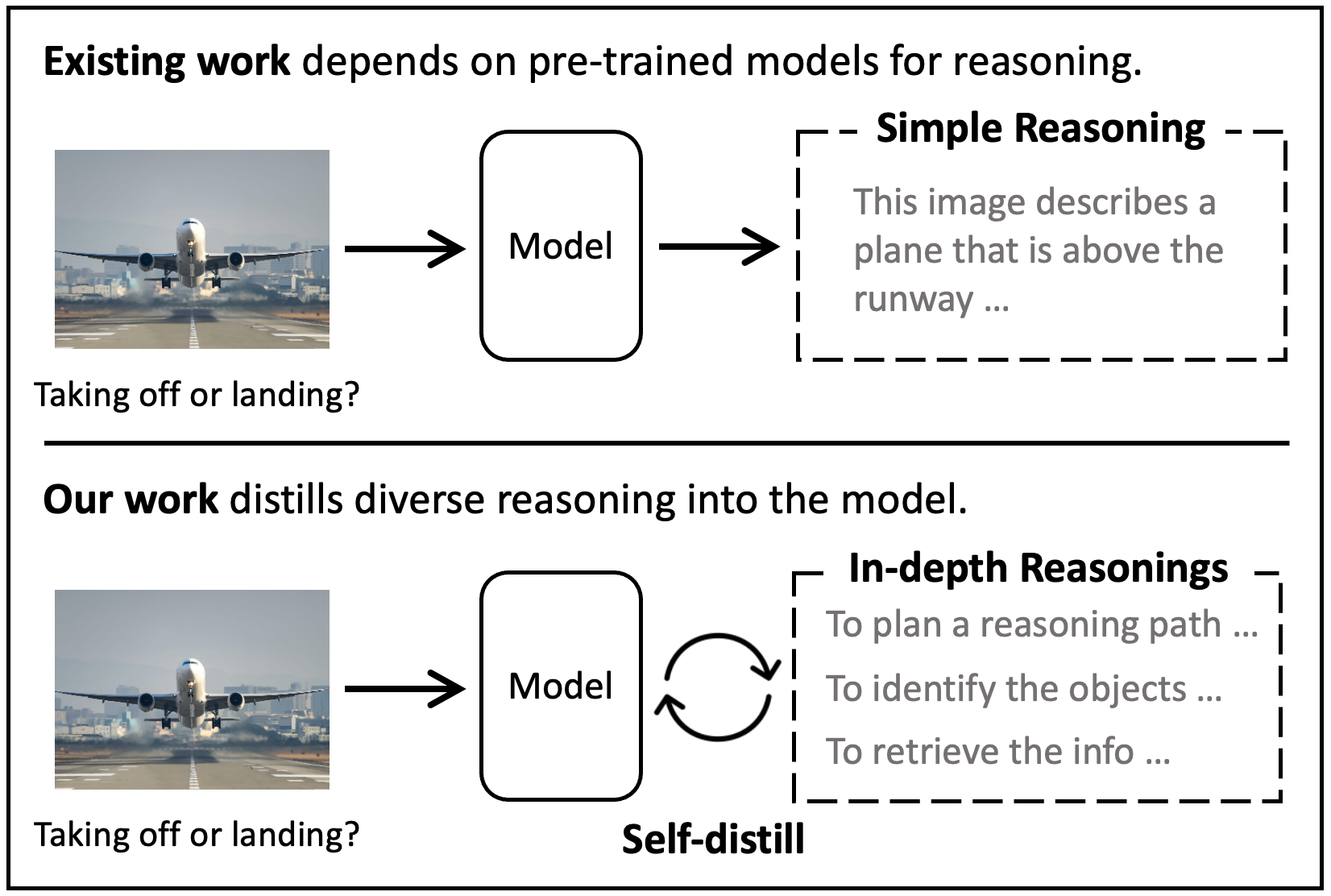}
    \caption{The comparison of existing work and our work. Instead of solely relying on reasoning capabilities of pre-trained models, our work proposes a self-distillation pipeline to integrate in-depth reasoning traces into the model.}

    \label{fig:framework}
\end{figure}


To address the limitations in cross-modal reasoning capabilities in VLMs, we propose a self-distillation framework to obtain reasoning traces and subsequently integrate them into the model.
The proposed framework first elicits comprehensive reasoning traces from the VLM with both two-step and diverse prompts, and then fine-tunes the VLM itself with the set of latent reasoning outputs. This approach ensures that sufficient reasoning traces are captured by the model. We leverage a diverse set of reasoning-invoking prompts to capture different in-depth reasoning needed for arriving at the answer. To facilitate effective self-distillation, we incorporate several critical designs into the VLM architecture: a representation intervention module that efficiently captures the reasoning traces, an ensemble weighting module that automatically adapts the diverse reasoning traces to the input data, and a skip-connection module to further bridge the gap between the text and image modalities. To summarize our main contributions:

\begin{itemize}[nosep,leftmargin=*]
\item We propose a self-distillation framework that obtains diverse in-depth reasoning traces and enhances reasoning capability of VLMs, addressing the limitations in reasoning capabilities dependent on pre-trained models.
\item We develop a teacher reasoning solicitation pipeline that uses prompt library to generate diverse in-context questions and utilizes a two-step reasoning procedure to obtain the reasoning-guided responses for student learning.
\item To allow the student to effectively capture the teacher reasoning traces, we add to the student model architecture several innovative components, including an intervention adapter, a cross-modal skip connection, and an ensemble learning algorithm.
\item Extensive experiments demonstrate the superiority of our method across five widely recognized VQA benchmarks, with performance improved from \conditionaltext{47.30\% to 53.85\%}{52.90\% to 59.10\%} for ANLS score and from 57.1\% to 62.67\% for accuracy compared to the baseline methods.
\end{itemize}

%% file: sections/05_related.tex
\section{Related Work}
\subsection{Multi-Modal LLMs}

Multi-modal LLMs or VLMs have gained popularity in recent years \cite{DBLP:journals/corr/abs-2404-01322, DBLP:journals/pami/ZhangHJL24, DBLP:conf/icml/RadfordKHRGASAM21, DBLP:conf/icml/0001LXH22, DBLP:conf/icml/0008LSH23, DBLP:conf/iclr/Zhu0SLE24}. Compared to previous ad-hoc multi-modal models \cite{DBLP:conf/icml/XuBKCCSZB15,DBLP:conf/aaai/00130SZ19}, multi-modal LLMs leverages the strong capabilities of pre-trained foundation models to offer greater generalizability to new domains, enabling few-shot and zero-shot learning. PaLM-E \cite{DBLP:conf/icml/DriessXSLCIWTVY23} introduced a novel approach by embedding visual information directly into a large language model, enabling multi-modal reasoning without the need for separate visual and textual encoders. LLaVA \cite{DBLP:conf/nips/LiuLWL23a} further developed this method by fine-tuning a vision encoder and language model end-to-end on image-text pairs, achieving strong performance across various vision-language tasks.
Commercial models like GPT-4V~\cite{DBLP:journals/corr/abs-2405-00732} and Gemini~\cite{DBLP:journals/corr/abs-2312-11805} have pushed the boundaries of the multi-modal LLMs' capabilities, demonstrating impressive zero-shot performance on a wide range of tasks.

Multi-modal LLMs can be adapted to new domains using fine-tuning methods. Since it's expensive to update all the model weights, recent studies focus on using the lightweight adapter modules~\cite{DBLP:journals/ijcv/GaoGZMFZLQ24, DBLP:conf/eccv/ZhangZFGLDQL22}, and parameter-efficient 
fine-tuning (PEFT). 
For example, CLIP-Adapter uses a lightweight adapter module that runs in parallel with the original CLIP model while freezing the CLIP weights during the fine-tuning stage.~\cite{DBLP:journals/ijcv/GaoGZMFZLQ24}.
Similarly, LLM PEFT methods such as LoRA \cite{DBLP:conf/iclr/HuSWALWWC22} reduce the fine-tuning overhead by freezing weights in the original model, while training additional low rank adapters. Although these fine-tuning approaches reduces the amount of weights that are updated, they still require high-quality annotated data, which is often unavailable.

Apart from fine-tuning, chain-of-thoughts (CoT) prompts have been used on LLM to elicit reasoning steps and improve performance ~\cite{DBLP:conf/nips/Wei0SBIXCLZ22}. Multi-modal LLMs can be prompted with CoT instructions as well to achieve better result ~\cite{DBLP:conf/cvpr/ZhouYL022, DBLP:conf/eccv/JiaTCCBHL22, DBLP:journals/corr/abs-2403-16999, DBLP:conf/nips/ZhangZ023, DBLP:conf/cvpr/ZhangQP0J24}. 
Our approach follows this paradigm by using dual-query prompts to elicit diverse reasoning paths and generate high quality dataset. The dataset is then used to train intervention modules that modify the LLM's representations. This method removed the need for high-quality annotations while also reduce the amount of weights that need update.


\subsection{Knowledge Distillation}
Knowledge distillation was introduced as a way to transfer knowledge from a teacher model to a student model~\cite{hinton2015distillingknowledgeneuralnetwork}. Through distillation, the student model can achieve a better performance-size trade-off. As a special case of knowledge distillation, self-distillation is a family of methods where teacher and student use identical model architecture. Early work in this area focused on convolutional neural networks, demonstrating improved performance, generalization, and efficiency~\cite{DBLP:conf/iccv/ZhangSGCBM19, DBLP:journals/pami/ZhangBM22, DBLP:journals/corr/abs-2206-08491}. Theoretical studies have shown that self-distillation amplifies regularization in Hilbert space and affects optimization dynamics, providing insights into its effectiveness~\cite{DBLP:conf/nips/MobahiFB20}.
Recent research has extended self-distillation to LLMs, leveraging the model's own generation to create dataset for self-improvement. By designing sophisticated data generation pipeline or loss functions, the distilled models show significant performance gains in various natural language understanding and generation tasks~\cite{ wang2022self, agarwal2024policy, DBLP:journals/corr/abs-2407-06023}. Our work aligns with the notion of self-distillation by using the same model as teacher and student, distilling dual-query embedding into single-query embedding. To the best of our knowledge, there is no existing work on applying self-distillation to the multi-modal LLM.

%% file: sections/02b_framework_hs.tex
\begin{figure*}[h!]
    \centering
            \includegraphics[width=\textwidth]{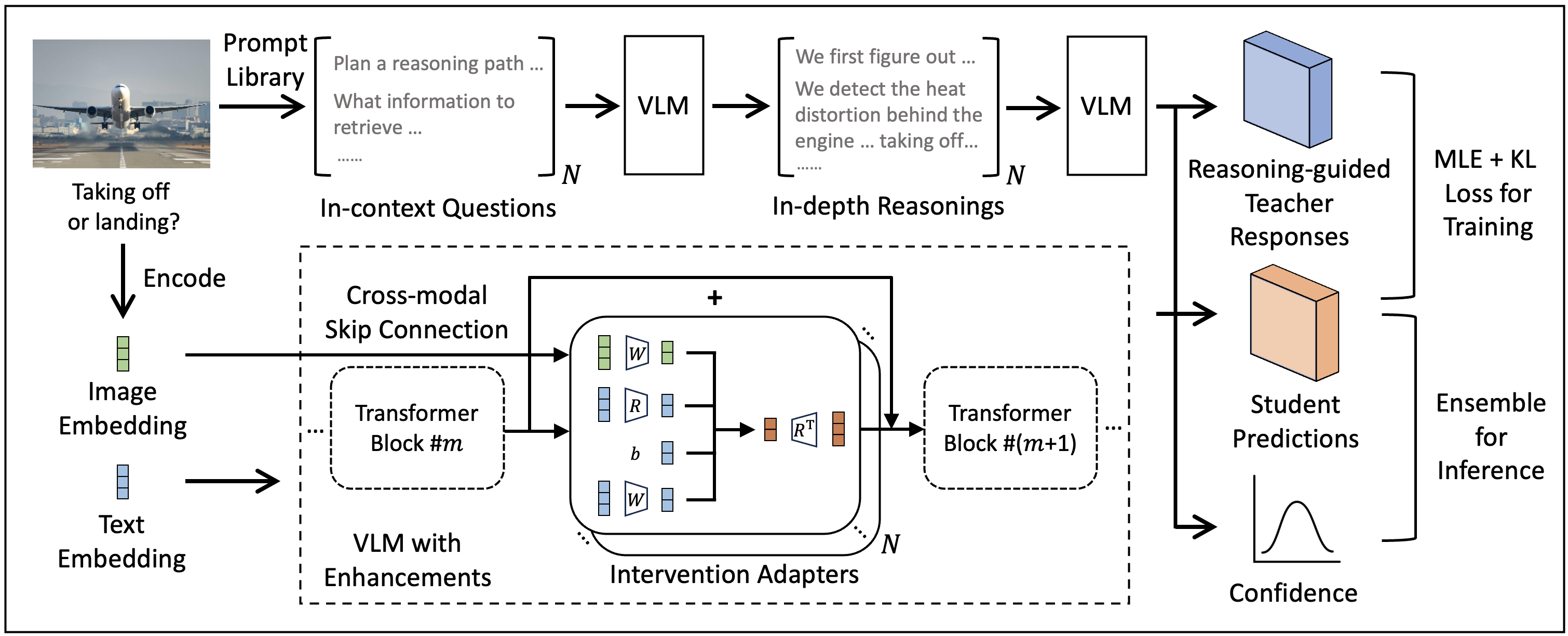}
    \caption{The overall framework. We first introduce a prompt library to generate $N$ in-context question pairs. These questions are then fed into the model in 2 consecutive steps to produce diverse reasoning-guided responses in parallel. Then, the obtained responses are utilized for self-distillation, enabling the model to better internalize the reasoning process. To enhance the model’s ability to capture reasoning effectively, we design and add several innovative components into the model architecture, including an intervention adapter for efficient parameter updates, a cross-modal skip connection to facilitate information exchange across different modalities, and an ensemble learning algorithm to synthesize reasoning derived from multiple in-context questions.
    }    
    \label{fig:framework}
\end{figure*}

\section{Methodology}
\label{sec:framework}
The overview of our proposed self-distillation framework is illustrated in Figure \ref{fig:framework}. The framework comprises two primary components: the reasoning trace generation pipeline and the enhanced VLM model architecture. The reasoning trace generation pipeline provides in-depth and diverse reasoning traces. These reasoning traces are then used to fine-tune the VLM, which has been enhanced with several key architectural changes. These changes include representation intervention modules, vision-language skip connection, and learnable ensemble weights. In the following subsections, we first describe the reasoning trace generation process, and then provide details about the architectural improvements made to the VLM.\looseness=-1

\subsection{Reasoning Trace Generation Pipeline}
The objective of this pipeline is to generate in-depth and diverse visual reasoning traces that can be used as strong supervision signals to fine-tune the VLM through self-distillation. To facilitate this, we expand upon the two-stage prompting process from \cite{DBLP:journals/tmlr/0001Z00KS24}, which consists of an intermediate analysis generation stage and an answer inference stage. Specifically, we introduce a diverse prompt library, where each prompt consists of two instructions applied to the VLM sequentially. These prompts focus on various two-step reasoning paths, such as information extraction, automatic planning, text layout analysis and visual concept alignment. An example is given in Figure \ref{fig:framework} and the full prompt library is given in Appendix.

Importantly, to enable effective self-distillation, we obtain hidden states from the second step, rather than the final responses, as the training signal. Formally, let $p_1^n$ and $p_2^n$ represent the $n$-th prompt in the library corresponding to the instructions for the two query steps. In the first step, the instruction $p_1^n$ is concatenated with the input $x$ (consisting of both the image and question) and passed through the VLM. We focus on VLMs consisting of a series of transformer decoder layers. Let $Output$ refer to the output layer that transforms the hidden states output by the decoder layer to the probability of the next token, and $Decode$ refer to the token decoding mechanism. We denote representations of the last hidden layer from the first and second steps as $h_1^n$ and $h_2^n$ respectively, the generated token sequence as $y_1^n$, $y_2^n$, and the decoded answer as $\hat{y_1^n}$, $\hat{y_2^n}$. Therefore, for the first query step, we obtain the following: \looseness=-1
\begin{align}
P(y_1^n|[p_1^n , x]) &= Output(h_1^n) = Output(f([p_1^n,x])), \\ \hat{y_1^n} &= Decode(P(y_1^n|[p_1^n , x]))
\end{align}

In the second query step, we concatenate the first-query answer $\hat{y_1^n}$, the input $x$, the prompt $p_2^n$, and obtain the generation probability:  \looseness=-1
\begin{align}
    P(y_2^n | [p_1^n,x, \hat{y_1^n}, p_2^n]) = Output(h_2^n) \\ =  Output(f([p_1^n,x,\hat{y_1^n}, p_2^n]))])), \nonumber \\
    \hat{y_2^n} = Decode(P(y_2^n | [p_1^n,x,\hat{y_1^n}, p_2^n]))
    \label{eq:query2}
\end{align}
Finally, the set of hidden states $\{h_2^n\}_{n=1}^N$ obtained through this pipeline-which encode rich contextual information and reasoning traces—serve as the ground truth signals used to fine-tune the VLM for improved reasoning capability. 

\subsection{VLM Architectural Enhancements for Self-Distillation}
Following our reasoning trace generation pipeline, we utilize the generated hidden states to enable single-step inference through self-distillation. As outlined below, we developed several key architectural enhancements to the original VLM to enable effective self-distillation. 

\subsubsection{Intervention Adapter}
Rather than training the entire VLM which could consist of billions of parameter, we adapt the representation fine-tuning (ReFT) \cite{DBLP:journal/corr/abs-2404-03592} approach originally proposed for LLMs to our VLM architecture for efficient training. The ReFT adapter modules consist of low-rank decomposed representation updates for the transformer hidden states. We insert them as parallel adaptation layers at the output of each LLM transformer block, as shown in Figure~\ref{fig:framework}, but not on the vision transformer branch.
This approach significantly reduces the number of parameters required for fine-tuning the VLM. 

Formally, we implement ReFT modules for VLM as intervention block $I_{ReFT}^n = B\times A \times h$, where $B \in \mathcal{R}^{r \times k}$ and $A \in \mathcal{R}^{k \times r}$, with $k$ representing the dimension of the hidden states and $r$ being the low rank, which is substantially smaller than $k$. In line with~\cite{DBLP:conf/iclr/HuSWALWWC22}, we initialize $A$ with a random Gaussian distribution and set $B$ as a zero matrix. Following~\cite{DBLP:journal/corr/abs-2404-03592}, we also integrate control over which positions in the input sequence these interventions occur since certain prefix/suffix often contain more significant or relevant information related to the task. It is observed that such approach achieves smaller gradients during training while reducing overall parameters. Specifically, we only modify the first and last $pos$ positions where $pos$ is a configurable hyperparameter and adjusted during training.


\subsubsection{Sell-Distillation Loss}
To train the intervention module $I^n$, we can use mean squared error (MSE) loss, the Kullback-Leibler (KL) divergence or a combination of the two as the learning objective. For simplicity, we denote $h$ as the hidden states of the base transformer module shared among all intervention modules. Since our intervention module $I^n$ runs parallel with the transformer layers, we can optimize the module only without updating the base transformer. To compute the MSE loss for the $n$-th intervention module, we define:  
\begin{align}
    \hat{h} &=I^n(h) + h \\
    \mathcal{L}^n &= MSE (h_2^n, \hat{h}) 
    \label{eq:loss_mse}    
\end{align}

For the KL-divergence, we define the following loss term:
\begin{align}
    \mathcal{L}_{kl} = KL (Output(h_2^n), Output(\hat{h}))
    \label{eq:loss_kl}    
\end{align}

These two loss formulations can be utilized independently or jointly to train the self-distilled model. We provide an ablation study in Section \ref{section: ablation_loss} on the effects of different formulations. \looseness=-1

Equations~\ref{eq:loss_mse} and~\ref{eq:loss_kl} are specific to one layer of the VLM for fine-tuning. However, we can generalize it to apply to multiple layers, by adding up the losses for different layers. In our implementation, we only intervene in the last 3 layers since the deeper layers encode the high-level vision-language features.




\subsubsection{Vision-Language Skip Connection}
We observe that VLM may struggle with low-level perceptual tasks such as object counting and object detection, which can serve as intermediate steps in multi-hop VQA tasks~\cite{DBLP:conf/cvpr/YuanWJC21}. To address this and better support the in-context learning with the visual reasoning complexity, we propose a cross-modal skip connection schema that embeds the early layer outputs of the vision encoder into the intervention module. This method leverages low-level vision features to enhance VLM's perceptual capabilities. Specifically, the intervention module is implemented as:
\begin{equation}
    I_{skip}^n = R^T (W h + b + A u)
\end{equation}
where $u$ is the feature extracted from the vision encoder, $A \in \mathcal{R}^{||u|| \times k}$ is the matrix for projecting the visual features into the subspace, and $R$ is the low-rank matrix which restores the hidden state dimensionality. The resulting sum of $I_{Reft}^n$ and $I_{skip}^n$ is then passed into the subsequent layer of the VLM.

\subsubsection{Ensemble Learning}
During training, distinct prompts create varied output hidden state for the same input image and questions, enabling us to train separate intervention adapters accordingly. After obtaining these trained representation modules, we introduce a learnable predictor to produce ensemble weighting vector $w\in R^N $. To train the predictor, we select the reasoning trace with the highest next-token confidence - determined by top-token probability - as the ground truth. The underlying assumption here is that higher skewness in token probability distribution indicates greater model confidence~\cite{DBLP:conf/icml/GuoPSW17, DBLP:conf/naacl/GengCWKNG24}. 


%% file: sections/04_experiments.tex
\section{Experiments}
\subsection{Experimental Setup}
\textbf{Datasets.} We evaluate our approach on five widely used VQA datasets, i.e., DocVQA~\cite{DBLP:conf/wacv/MathewKJ21}, InfographicVQA~\cite{DBLP:conf/wacv/MathewBTKVJ22}, MMMU~\cite{yue2024mmmu}, ChartQA~\cite{DBLP:conf/acl/MasryLTJH22}, and FigureQA~\cite{DBLP:conf/iclr/KahouMAKTB18}. In particular, DocVQA contains 30K questions and 5K images related to documents. 
\conditionaltext{We evaluate the models on the test split of DocVQA instead of the valid split used in ~\cite{DBLP:conf/nips/LiuLWL23a}. }{We evaluate the models on the valid split of DocVQA~\cite{DBLP:conf/nips/LiuLWL23a}. }
InfographicVQA (InfogVQA) contains 5K infographic images and 30K related questions. This dataset challenges the model's ability to comprehend and extract information from complex visual structures like infographics. ChartQA contains 9.6K human-written questions and 23.1K questions on the charts. 
ChartQA and FigureQA require the models to understand the data visualization and extract the data insights correctly. MMMU contains 11.5K collected multi-modal questions from the exams across 30 domains.

\conditionaltext{\input{tables/quantitative} }{\input{tables/quantitative_docvqa_valid} }

\noindent
\textbf{Metrics.}
We evaluate the models based on the default metrics provided by the different datasets. For DocVQA and InfographicVQA, we use the average normalized Levenshtein similarity (ANLS) score~\cite{DBLP:conf/naacl/AndreasRDK16}, which compares the generated sentence with the reference answers using Levenshtein distance, avoiding issues with slight variations compared to exact match. For ChartQA and FigureQA, accuracy is used for evaluation. For MMMU, we utilize micro-averaged accuracy, applying regular expressions to extract the response from the answer, including immediate reasoning steps.


\noindent
\textbf{Hardwares.}
All experiments were conducted on a cloud server with 4 Nvidia A100 80G GPUs.

\begin{figure}
    \centering
    \includegraphics[width=0.42\textwidth]{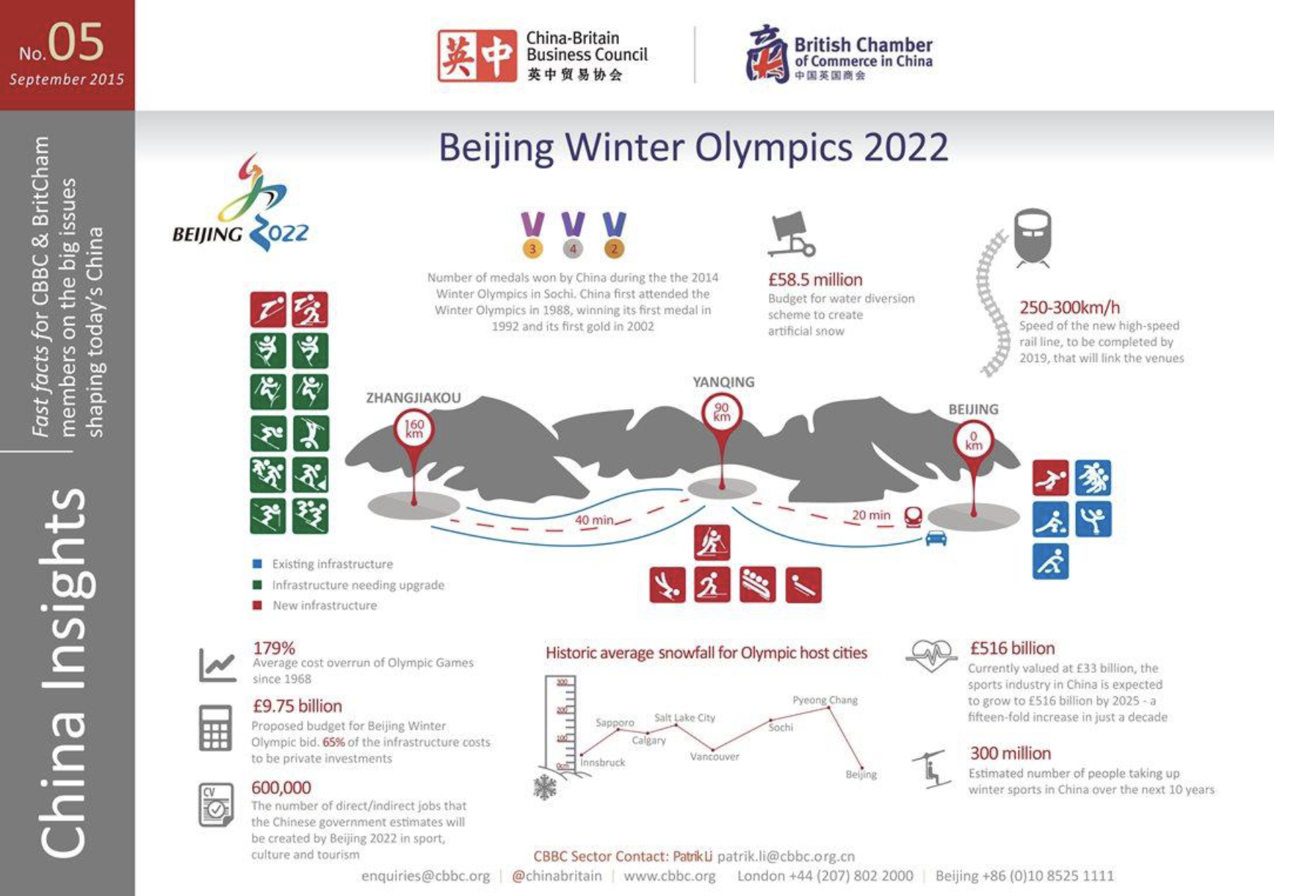}
    \caption{An example figure in InfographicVQA dataset test split. The question is ``\textit{what is the total number of infrastructures in Zhangjiakou that need to be upgraded?}''}
    \label{fig:exp-example1}
\end{figure}

\input{tables/category_mmmu}
\input{tables/category_chartqa}

\conditionaltext{\input{tables/ablation_skipc}}{\input{tables/ablation_skipc_docvqa_valid}}
\conditionaltext{\input{tables/ablation_loss}}{\input{tables/ablation_loss_docvqa_valid}}

\conditionaltext{\input{tables/ablation_prompt}}{\input{tables/ablation_prompt_docvqa_valid}}
\conditionaltext{\input{tables/reft_vs_lora}}{\input{tables/reft_vs_lora_docvqa_valid}}

\subsection{Performance Comparison}
All experiments were conducted using LLaVA-1.6 with Mistral 7B. For performance evaluation, we selected three baseline methods, including basic prompting, visual chain-of-thought (CoT) prompting~\cite{DBLP:journals/tmlr/0001Z00KS24}, and fine-tuning using Low-Rank Adaptation (LoRA). We present our experiment result in Table~\ref{tab:quantitative}. The result indicates a consistent performance increase on all five datasets. Below, we present a more detailed analysis.

\noindent
\textbf{MMMU.} This dataset contains the difficult VQA problems requiring connecting multiple steps of reasoning and applying the domain-specific knowledge in science and engineering fields. These problems also present challenges in interpreting the abstract visual content such as diagrams, charts, and mathematical expressions alongside relevant contextual information. In Table~\ref{tab:quantitative}, our approach outperforms all three baselines.  

\noindent
\textbf{DocVQA and InfographicVQA (InfogVQA).}
Our approach outperforms all three baselines on DocVQA and InfographicVQA datasets. 
For DocVQA, the performance increase (3.5\%) indicates our approach's effectiveness in understanding document structure and text-visual relationships. The larger gain in InfographicVQA (9.6\%) suggests enhanced spatial reasoning capabilities for layout analysis and multi-hop reasoning, which are crucial for interpreting illustrations.
For example in Figure~\ref{fig:exp-example1}, our approach can successfully solve the question \textit{what is the total number of infrastructures in Zhangjiakou are needing upgrade} and provide the answer $2$. We find that the baseline struggles, while the two-stage prompting pipeline can first notice that the question is focused on Zhangjiakou instead of all sites (Beijing and Zhangjiakou). In this case, the decomposed query and subsequently the self-distilled model can help the model to identify the key elements in the image and reason with the identified information.  

\noindent
\textbf{ChartQA and FigureQA.}
ChartQA and FigureQA focus on chart and visualization-based question-answering. These tasks require both low-level perceptual abilities for data parsing and spatial reasoning capabilities for interpreting various chart elements. Our approach achieved better accuracy on these three datasets.
Visualization data differs significantly from natural images that dominate multi-modal LLM pre-training corpora. The performance improvement in those two datasets indicate our approach's generalizability and effectiveness in domain transfer when the labeled data is scarce. 

\subsection{Analysis and Ablation Studies}

\noindent

\textbf{Category-Wise Analysis.}
We conduct the category-wise analysis on ChartQA and MMMU datasets. 
ChartQA defines the questions by four types, i.e., \textit{data retrieval, visual, compositional, and visual\&compositional tasks}. We sample and self-annotate 200 samples from ChartQA.
We present the result in Figure~\ref{fig:category_chartqa}. In particular, we find our approach outperformed the baseline the mostly on the more challenging \textit{compositional and visual\&compositional} task (where the baseline performance is the lowest).

For MMMU dataset, we include 32 different domains, which enable us to evaluate our approach under different domains. The dataset further summarizes the domains into sub-categories, i.e., \textit{ business, science, health\&medicine, and technology\&engineering}.
We present the category-wise experiment result in Table~\ref{tab:category_mmmu}.
The results indicate that our approach shows an especially promising result in the domains of \textit{business} and \textit{science}, which require intensive reasoning process~\cite{yue2024mmmu}. 
Moreover, we notice that our approach does not exhibit significantly better performance on \textit{humanity\&social science} and \textit{technology\&engineering}. We attribute this to the challenges posed by commonsense and domain-specific knowledge. 

\textbf{Ablation: Vision-Language Skip Connection.}
We verify the effectiveness of our proposed vision-language skip connection by comparing our full approach with a version that excludes the skip connection module. The results, shown in Table~\ref{tab:ablative_skipc}, indicate that the skip connection improves model performance, suggesting that low-level visual features are crucial for the VQA tasks and it is beneficial to include the low-level features as input to the intervention module.

\textbf{Ablation: Number of Ensemble Members.}
We conduct an ablative study for finding the optimal number of ensemble members and present the result in Figure~\ref{fig:ablation_n}.
We list the prompts we used in the Appendix. 
Our findings suggest that $K=4$ (4 ensemble members) provides a good balance between the number of members and performance. Although the $K=8$ model achieves similar performance, the additional prompts introduce extra training overhead.




\textbf{Ablation: Loss Function for Distillation.}
\label{section: ablation_loss}
We analyze different loss functions used for self-distillation, i.e., MSE, KL divergence, and their combination (MSE + KL). The KL divergence loss is computed in the token probability space rather than the representation space. We present the results in Table~\ref{tab:ablation_loss}, which shows that the combination of MSE and KL divergence yields the optimal results.

\textbf{Ablation: Adapters Used in Model.}
\label{section: reft_vs_lora}
We analyze the performance of applying ReFT intervention or LoRA adapter into our self-distillation framework design. Compared to LoRA adapter which is widely used into the fine-tuning , the intervention adapter yields the better performance in our self-distillation framework in Table~\ref{tab:reft_vs_lora}.

%% file: tables/quantitative.tex
\begin{table*}[]
\centering
\resizebox{.6\linewidth}{!}{
\begin{tabular}{lccccc}
\toprule
Method & MMMU & DocVQA & InfogVQA & ChartQA  & FigureQA \\ \midrule
Prompting                & 36.1 & 60.0   & 34.6     & 56.1     & 79.1     \\ \cmidrule{1-6} 
Visual CoT Prompting                & 38.7 &  62.5   & 38.0     & 56.6     & 81.5     \\ \cmidrule{1-6}
Fine-tuning              & 41.9 & 60.1   & 41.2     & 58.7     & 83.1     \\ \cmidrule{1-6}
Ours    & \textbf{42.6} & \textbf{63.5} & \textbf{44.2} & \textbf{61.9} & \textbf{83.5} \\ \bottomrule
\end{tabular}
}
\caption{Performance Comparison. Our proposed self-distillation approach consistently outperforms all three baseline approaches across all five VQA datasets.}
\label{tab:quantitative}
\end{table*}

%% file: tables/quantitative_docvqa_valid.tex
\begin{table*}[]
\centering
\resizebox{.6\linewidth}{!}{
\begin{tabular}{lccccc}
\toprule
Method & MMMU & DocVQA & InfogVQA & ChartQA  & FigureQA \\ \midrule
Prompting                & 36.1 & 71.2  & 34.6     & 56.1     & 79.1     \\ \cmidrule{1-6} 
Visual CoT Prompting                & 38.7 &  72.4   & 38.0     & 56.6     & 81.5     \\ \cmidrule{1-6}
Fine-tuning              & 41.9 & 71.4   & 41.2     & 58.7     & 83.1     \\ \cmidrule{1-6}
Ours    & \textbf{42.6} & \textbf{74.0} & \textbf{44.2} & \textbf{61.9} & \textbf{83.5} \\ \bottomrule
\end{tabular}
}
\caption{Performance Comparison. Our proposed self-distillation approach consistently outperforms all three baseline approaches across all five VQA datasets.}
\label{tab:quantitative}
\end{table*}

%% file: tables/category_mmmu.tex
\begin{table*}[t]
\centering
\resizebox{.55\linewidth}{!}{
\begin{tabular}{lcccc}
\toprule
Method          & \textit{Business} & \textit{Science} & \textit{Health \& Medicine} & \textit{Tech \& Eng.} \\ \midrule
Prompting      & 29.4 & 28.8  & 39.2 & 28.7 \\ \midrule
Fine-tuning     & 32.1 & 32.5 & 42.1 & 30.6 \\ \midrule
Ours & \textbf{39.2} & \textbf{36.4} & \textbf{47.8} & \textbf{33.4} \\ \bottomrule
\end{tabular}}
\caption{Category-wise analysis result on MMMU dataset. MMMU dataset divides the data into sub-categories including \textit{Business}, \textit{Science}, \textit{Health\&Medicine}, and \textit{Technology\&Engineering} etc. We compare our approach with the baseline prompting and LoRA-based fine-tuning methods.}
\label{tab:category_mmmu}

\end{table*}

%% file: tables/category_chartqa.tex


\begin{figure}[t]
    \centering
    \includegraphics[width=0.5\textwidth]{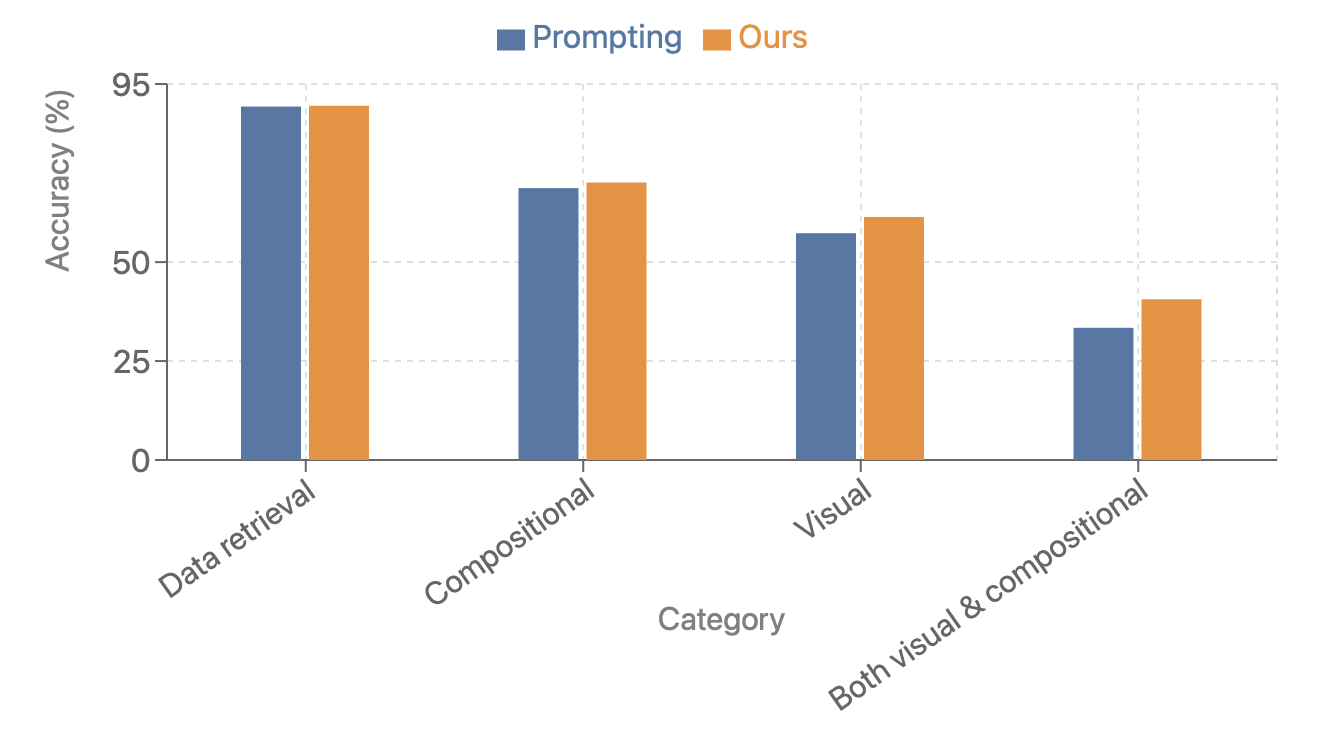}
    \caption{Category-wise analysis on ChartQA dataset. The data comprises of the four different question categories, i.e., \textit{data retrieval}, \textit{compositional}, \textit{visual} and \textit{both visual \& compositional}.}
    \label{fig:category_chartqa}
\end{figure}

%% file: tables/ablation_skipc.tex
\begin{table}[]
\centering
\resizebox{.42\textwidth}{!}{
\begin{tabular}{cccc}
\toprule
Model & Skip C. & DocVQA & InfogVQA \\ \midrule
 \multirow{2.5}{*}{Ours}  & Y               &   \textbf{63.5}      &     \textbf{44.2}      \\ \cmidrule{2-4}
  & N               &    61.3    &  40.5         \\ \bottomrule
\end{tabular}}

\caption{Ablative study for the skip-connection mechanism. Skip C. refers to whether the condition contains the skip connection method in the intervention module.}
\label{tab:ablative_skipc}

\end{table}

%% file: tables/ablation_skipc_docvqa_valid.tex
\begin{table}[]
\centering
\resizebox{.7\columnwidth}{!}{
\begin{tabular}{cccc}
\toprule
Model & Skip C. & DocVQA & InfogVQA \\ \midrule
 \multirow{2.5}{*}{Ours}  & Y               &   \textbf{74.0}      &     \textbf{44.2}      \\ \cmidrule{2-4}
  & N               &    71.8    &  40.5         \\ \bottomrule
\end{tabular}}

\caption{Ablative study for the skip-connection mechanism. Skip C. refers to whether the condition contains the skip connection method in the intervention module.}
\label{tab:ablative_skipc}

\end{table}

%% file: tables/ablation_loss.tex
\begin{table*}[]
\centering
\resizebox{.5\linewidth}{!}{
\begin{tabular}{lccc}
\toprule
Training          & Loss        & DocVQA & InfographicVQA \\ \midrule
\multirow{3.8}{*}{Self-Distillation} & MSE         & \textbf{63.5}    & 42.2  \\ \cmidrule{2-4}
 & KL          & 55.4      & 38.1  \\ \cmidrule{2-4}
 & KL\&MSE     & \textbf{63.5}      & \textbf{44.2}  \\ \bottomrule
\end{tabular}
}
\caption{Abaltive study for the loss used in self-distillation. The combination of KL and MLE yields the best performance.}
\label{tab:ablation_loss}

\end{table*}

%% file: tables/ablation_loss_docvqa_valid.tex
\begin{table*}[]
\centering
\resizebox{.5\linewidth}{!}{
\begin{tabular}{lccc}
\toprule
Training          & Loss        & DocVQA & InfographicVQA \\ \midrule
\multirow{3.8}{*}{Self-Distillation} & MSE         & \textbf{73.9}    & 42.2  \\ \cmidrule{2-4}
 & KL          & 68.3      & 38.1  \\ \cmidrule{2-4}
 & KL\&MSE     & \textbf{74.0}      & \textbf{44.2}  \\ \bottomrule
\end{tabular}
}
\caption{Abaltive study for the loss used in self-distillation. The combination of KL and MLE yields the best performance.}
\label{tab:ablation_loss}

\end{table*}

%% file: tables/ablation_prompt.tex

\begin{figure}
    \centering
    \includegraphics[width=0.50\textwidth]{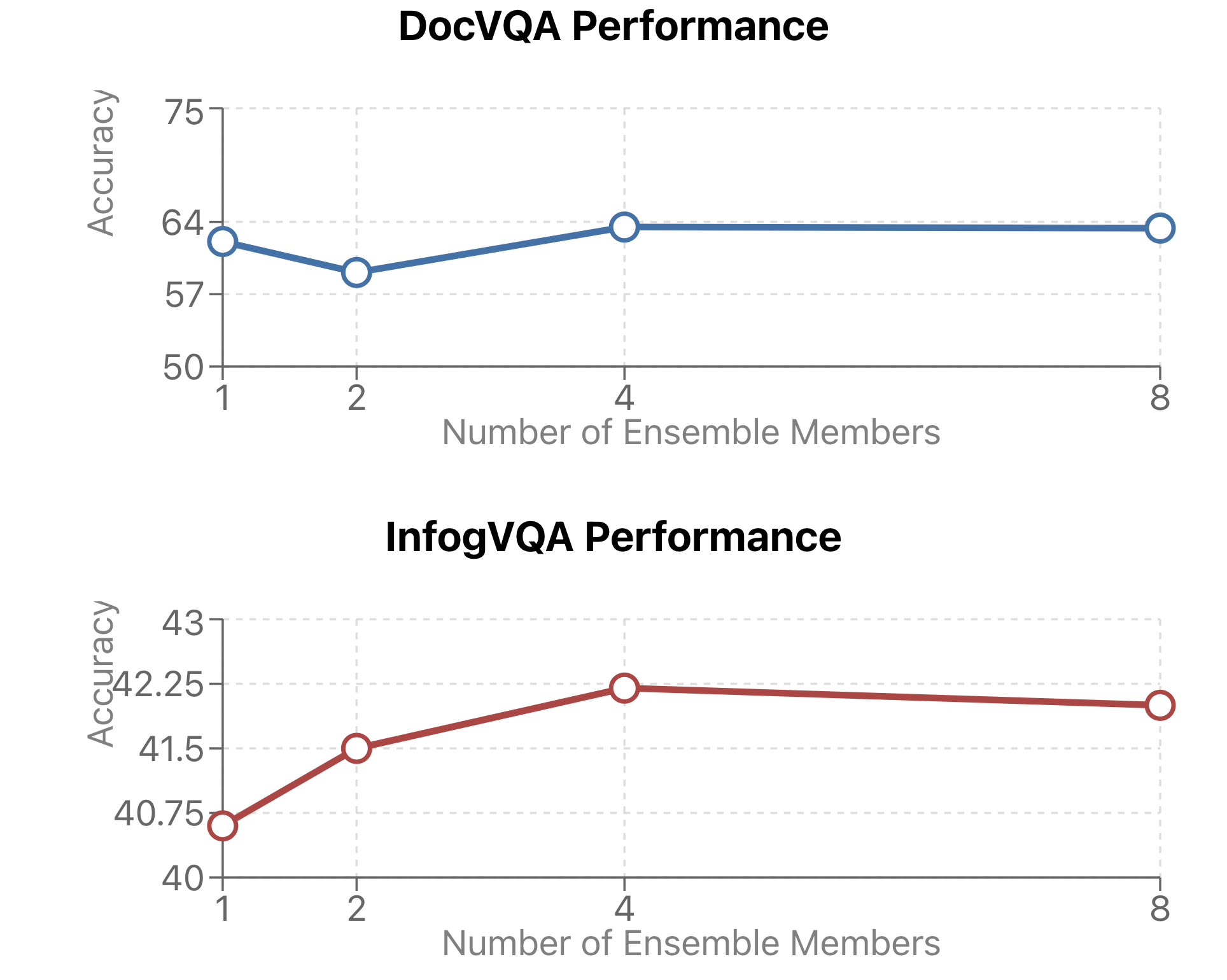}
    \caption{Effect of number of ensemble members in the self-distillation framework on DocVQA (top) and InfographicVQA (bottom) results.}
    \label{fig:ablation_n}
\end{figure}

%% file: tables/ablation_prompt_docvqa_valid.tex
\begin{figure}
    \centering
    \includegraphics[width=\linewidth]{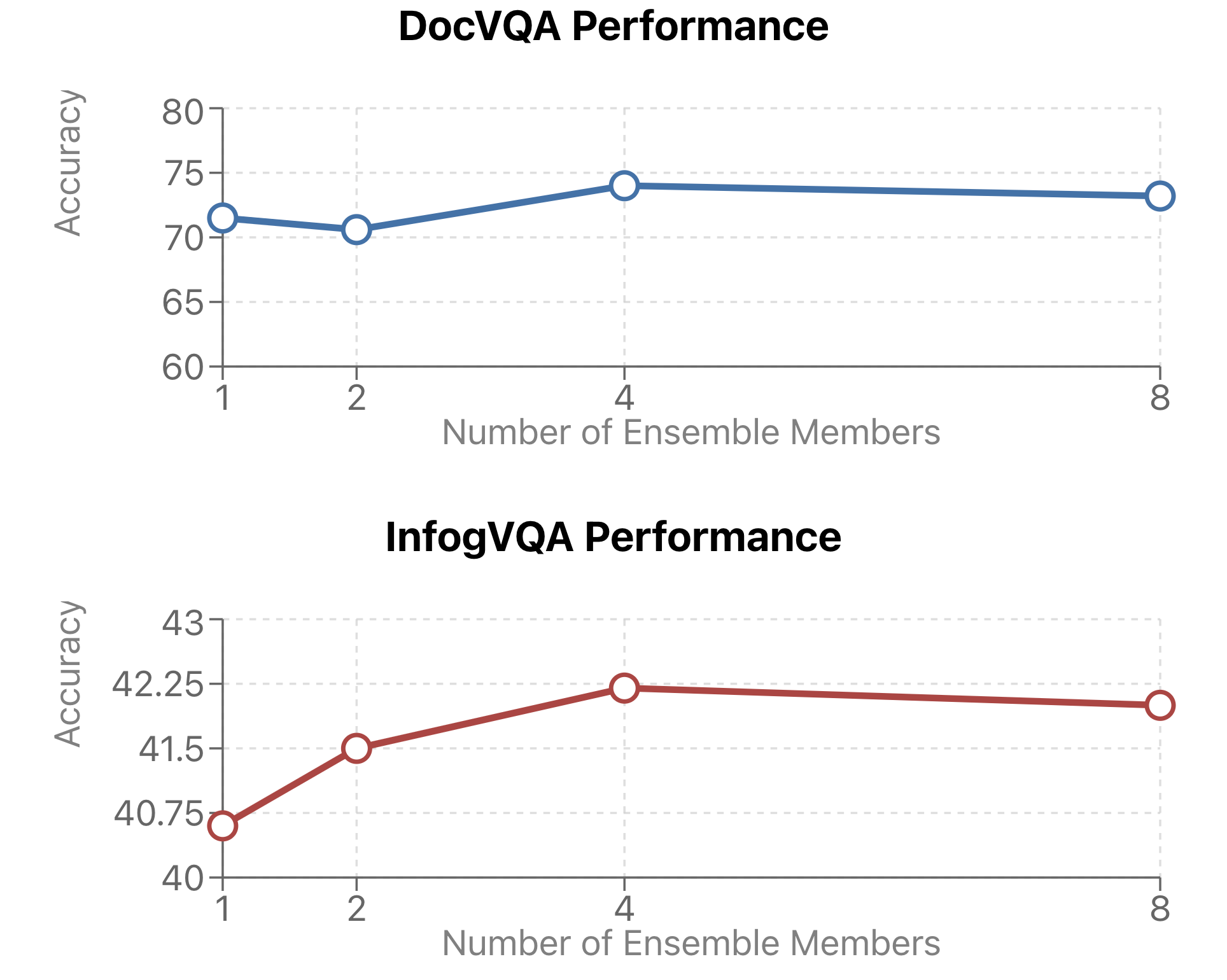}
    \caption{Effect of number of ensemble members in the self-distillation framework on DocVQA (top) and InfographicVQA (bottom) results.}
    \label{fig:ablation_n}
\end{figure}

%% file: tables/reft_vs_lora.tex
\begin{table*}[t]
\centering
\resizebox{.6\linewidth}{!}{
\begin{tabular}{lccccc}
\toprule
Adapter & MMMU & DocVQA & InfogVQA & ChartQA  & FigureQA \\ \midrule
Intervention        & \textbf{39.6} & \textbf{62.1}   & \textbf{40.6}     & \textbf{60.1}     & \textbf{83.1}     \\ \cmidrule{1-6}
LoRA        & 38.7 & 58.6   & 40.0     & 54.8     & 81.1     \\ \bottomrule
\end{tabular}
}
\caption{Ablation study for using intervention adapter or LoRA adapter in our self-distillation module}
\label{tab:reft_vs_lora}
\end{table*}

%% file: tables/reft_vs_lora_docvqa_valid.tex
\begin{table*}[t]
\centering
\resizebox{.6\linewidth}{!}{
\begin{tabular}{lccccc}
\toprule
Adapter & MMMU & DocVQA & InfogVQA & ChartQA  & FigureQA \\ \midrule
Intervention        & \textbf{39.6} & \textbf{72.6}   & \textbf{40.6}     & \textbf{60.1}     & \textbf{83.1}     \\ \cmidrule{1-6}
LoRA        & 38.7 & 71.5   & 40.0     & 54.8     & 81.1     \\ \bottomrule
\end{tabular}
}
\caption{Ablation study for using intervention adapter or LoRA adapter in our self-distillation module}
\label{tab:reft_vs_lora}
\end{table*}

%% file: sections/06_limitations.tex
\section{Limitation and Discussion}


\noindent
One limitation of our approach is the dependency on manually crafted prompt texts. While our experimental results indicate that ensemble performance stabilizes with four or more members, the effectiveness of our method still hinges on high-quality, chain-of-thought, dual-query prompts. In practical scenarios, expert-level prompt design may not always be feasible. Therefore, exploring the efficacy of our approach using less structured, lower-quality prompts would help delineate its practical boundaries and enhance generalizability.

For further advancements in our self-distillation framework, we propose integrating more sophisticated pipelines for generating reasoning traces. A natural direction is employing multi-turn queries (more than two turns) within the ensemble to derive final answers. Although our preliminary experiments indicated limited effectiveness relative to the increased inference overhead for queries exceeding two turns, we believe other methods of leveraging multi-turn reasoning may prove beneficial. We plan to investigate these alternative strategies in future work.

\noindent

%% file: sections/07_conclusion.tex
\section{Conclusion}
In this paper, we introduced a novel self-distillation framework designed to enhance the reasoning capabilities of Vision-Language Models (VLMs). Our method generates explicit reasoning-guided samples and leverages them to fine-tune the VLMs, thus enabling the models to internalize diverse reasoning processes. We improved the underlying VLM architecture to facilitate effective self-distillation with key components such as representation intervention adapters, cross-modal skip connection, and ensemble weighting module. We evaluate our approach using LLaVA-1.6 on five widely used VQA datasets~\cite{DBLP:journals/corr/abs-2203-12119}. Results demonstrate that our method significantly improves the baseline performances. In summary, our approach offers a flexible method for empowering the reasoning capabilities for VLMs while maintaining inference efficiency and model compactness. 

%% file: supp/prompt.tex
\newpage

\section{Appendix Prompt}
\label{section:appendix_prompt}
We list the prompts we use for an ensemble of different prompts.

\subsection{Prompt 1}
The following prompt decomposes the image first based on the image and uses the extracted information to solve the question.
\prompttext{
[INST] Look at the image:
1. What objects/regions are important?
2. Which parts relate to the question?
[/INST]
[INST] User:
{images}
Question: \{question\}
[/INST]
Analysis:
}
\prompttext{
[INST]Given your analysis above, provide the final answer to the question.[/INST]
[INST] USER: \{images\} [/INST]
[INST] Question: \{question\} [/INST]
[INST] \{query1\_answer\} [/INST]
Answer:}

\subsection{Prompt 2}
The following prompt does dual-stage reasoning by analyzing regions and words before providing answers.

\prompttext{
[INST] USER: {images}
Based on the image, Question: 
{question} 
Please first plan a reasoning path without filling in the data. The reasoning path should reflect the path toward the final answer based on both the image and text questions. Please use the structural knowledge triplet to represent the analysis.
[/INST]
Reasoning path:
}

\prompttext{
[INST] USER: {images}
Based on the image and planned reasoning path from the image:
{query1\_answer}
please answer the question briefly
{question}
[/INST]
[INST] Assistant: the answer is [/INST]
Answer:
}
\subsection{Prompt 3}
The following prompt does text-layout analysis by extracting text content and spatial information before addressing questions.

\prompttext{
[INST] USER: \{images\}
Based on the image, please:
1. Extract all visible text from the image
2. Describe the layout and positioning of text/elements
3. Note any tables, lists, or structured content.
Do not answer the question yet.
Question: \{question\}
[/INST]
Layout Analysis:
}

\prompttext{
[INST] USER: \{images\}
Using the extracted text and layout information:
{query1\_answer}
Please answer the question:
{question}
[/INST]
[INST] Assistant: The answer is [/INST]
Answer:
}

\subsection{Prompt 4}
The following prompt does visualization understanding by classifying chart types and data characteristics before answering chart-related questions.

\prompttext{
[INST] USER: \{images\}
Based on the image, please analyze:
1. What type of visualization is this (e.g., bar chart, line plot, scatter plot)?
2. What data types are represented (e.g., categorical, numerical, temporal)?
3. Identify key visual components (axes, legends, labels)'
Question: \{question\}
[/INST]
Analysis:
}

\prompttext{
[INST] USER: \{images\}
Using the visualization analysis:
\{query1\_answer\}
Please answer the visualization-related question:
\{question\}
[/INST]
[INST] Assistant: The answer is [/INST]
Answer:
}

\subsection{Prompt 5}
The following prompt does attention-based reasoning by identifying regions of interest before focusing on these areas for answering.

\prompttext{
[INST] USER: \{images\}
Based on the image and question: \{question\}
Please first:
1. Identify the specific regions of interest in the image
2. Explain why these regions are crucial for the question
3. Note any relationships between different regions
Do not answer the question yet, only identify relevant regions.
[/INST]
Region Analysis:
}

\prompttext{
[INST] USER: \{images\}
Using the identified regions of interest:
\{query1\_answer\}
Now focus on these regions to answer:
\{question\}
[/INST]
[INST] Assistant: Based on the identified regions, the answer is [/INST]
Answer:
}

\subsection{Prompt 6}
The following prompt does mathematical problem-solving by identifying mathematical elements and planning calculation steps before executing solutions.

\prompttext{
[INST] USER: \{images\}
For this math question: \{question\}
Please analyze:
1. What mathematical elements are visible (numbers, equations, graphs)?
2. What mathematical operations/concepts are needed?
3. Break down the problem into calculation steps

Do not solve yet, only outline the mathematical approach.
[/INST]
Math Analysis:
}

\prompttext{
[INST] USER: \{images\}
Using the mathematical analysis:
\{query1\_answer\}
Now solve step by step:
\{question\}
[/INST]
[INST] Assistant: Following the steps above, the answer is [/INST]
Answer:
}

\subsection{Prompt 7}
The following prompt does scientific reasoning by extracting scientific information and required knowledge before applying concepts to answers.

\prompttext{
[INST] USER: \{images\}
For this science question: \{question\}
Please analyze:
1. What scientific elements/data are shown in the image?
2. What scientific concepts are involved?
3. What external scientific knowledge is needed to answer this?
   - Required theories/principles
   - Relevant formulas/relationships
   - Key scientific terminology
Do not answer yet, only analyze information and knowledge needs.
[/INST]
Science Analysis:
}

\prompttext{
[INST] USER: \{images\}
Based on:
1. Image information: \{query1\_answer\}
2. Required scientific knowledge above
Please solve:
\{question\}
[/INST]
[INST] Assistant: Applying the scientific concepts, the answer is [/INST]
Answer:
}

\subsection{Prompt 8}
The following prompt does concept alignment by matching key concepts between questions and images before generating concept-grounded answers.

\prompttext{
[INST] USER: \{images\}
Based on the image, Question: 
\{question\}
Please analyze:
1. What are the key concepts present in both question and image?
2. Where exactly are these concepts shown in the image?
3. How does the overall scene context support these concepts?
Do not answer yet, only provide concept analysis.
[/INST]
Concept Analysis:
}

\prompttext{
[INST] USER: \{images\}
Based on the concept analysis:
\{query1\_answer\}
Please answer concisely:
\{question\}
[/INST]
[INST] Assistant: Based on the identified concepts, the answer is [/INST]
Answer:
}